\documentclass[letterpaper, 10 pt, conference]{ieeeconf}

\IEEEoverridecommandlockouts                              

\usepackage{graphicx} 
\usepackage{amsmath} 

\title{\LARGE \bf
Extending Multi-Object Tracking systems to better exploit appearance and 3D information
}

\author{Kanchana Ranasinghe$^{*1}$, Sahan Liyanaarachchi$^{*1}$, Harsha Ranasinghe$^{*1}$ and Mayuka Jayawardhana$^{*1}$
\thanks{$^{*}$Authors contributed equally.}
\thanks{$^{1}$Department of Electronic and Telecommunication Engineering, University of Moratuwa}}

\begin{document}

\maketitle
\thispagestyle{empty}
\pagestyle{empty}


\begin{figure*}[t]
\vspace{-0.7cm}
  \centering
  \includegraphics[ width=\textwidth]{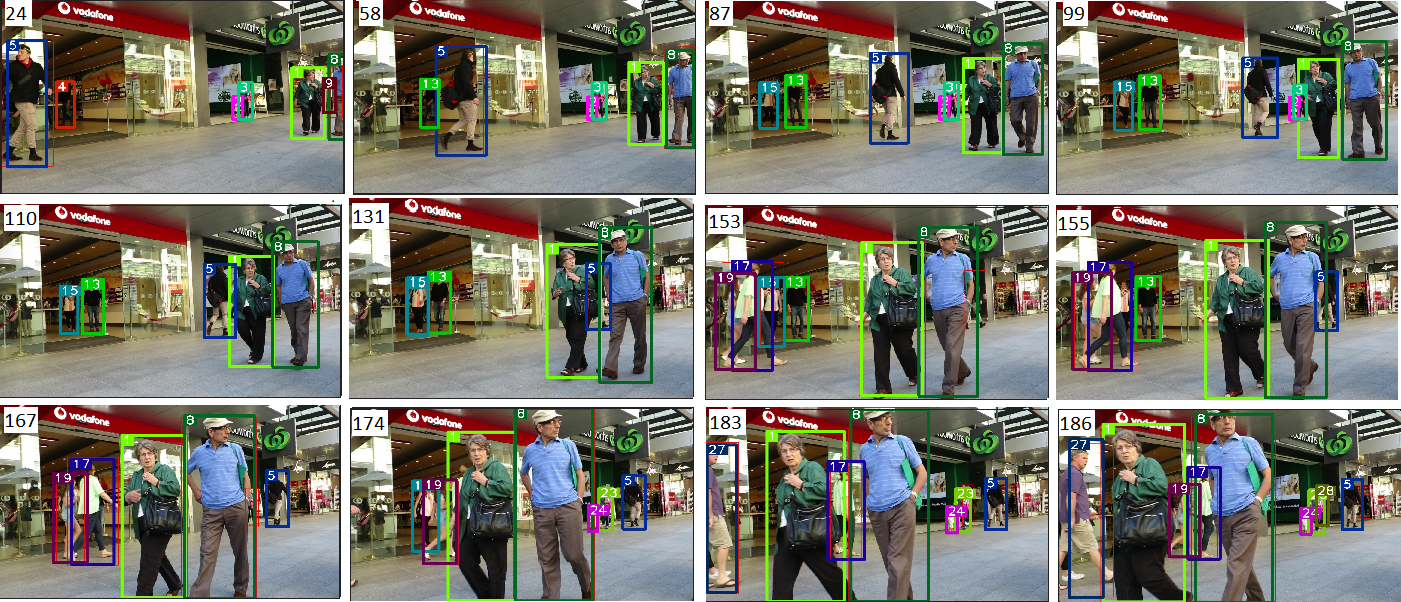}
  \vspace{-0.5cm}
  \caption{In the above diagram the numbers on the top left represent the frame number with detection bounding boxes in red and track bounding boxes in other colors. The above diagram depicts the occlusion handling done by our system. It can be seen that track 5(blue) observed in frame 24 is well tracked through out even after been occluded, given that the maximum occlusion duration is less than 30 frames. Additionally tracks 17 and 19 (blue and purple) are reidentified in frame 183 after been occluded. This is because, we control how much the occluding image’s features can affect the track’s template when the track is partially occluded\vspace{-0.4cm}}
  \label{fig:occlusions}
\vspace{-0.0cm}
\end{figure*}

\begin{abstract}

Tracking multiple objects in real time is essential for a variety of real-world applications, with self-driving industry being at the foremost. This work involves exploiting temporally varying appearance and motion information for tracking. Siamese networks have recently become highly successful at appearance based single object tracking \cite{DeepSiam:SiamFC, DeepSiam:siammask} and Recurrent Neural Networks have started dominating both motion and appearance based tracking \cite{DeepSiam:kalman, DeepSiam:multitarget, DeepSiam:deeptracking}. Our work focuses on combining Siamese networks and RNNs to exploit appearance and motion information respectively to build a joint system capable of real time multi-object tracking. We further explore heuristics based constraints for tracking in the Bird’s Eye View Space for efficiently exploiting 3D information as a constrained optimization problem for track prediction.

\end{abstract}

\section{Introduction}

Multi-object tracking has been a critical and unavoidable problem even at the level of cutting edge technology. State of the art multi-object tracking systems are computationally heavy for the end devices whereas real time tracking systems are performing at the expense of accuracy and even highly accurate systems \cite{DeepSiam:deepSort} make errors in general and edge cases like occlusions, ego-motion, crossovers and rapid/random movements. The occlusions build up heavy risks in the automobile industry that looks forward for self driving cars. The inability to predict a pedestrian crossing behind the vehicle that slowed down on the side front or the incapability of the tracker to distinctly identify two persons at a point of crisscrossing can lead to critical issues in the field that is chaotic and requires memory other than detection for producing near intelligent results.
\par Most of the current systems are based on tracking through detection where the extensive development of efficiency and accuracy in state-of-the-art frame level detectors is used. The novel idea is to incorporate the temporal aspects into the algorithm due to the seemingly simple fact that objects do not disappear and should follow a time dependent progression within frame sequence given a satisfactory sampling rate in the sequence. Moreover, the improvement of depth sensation from both monocular and binocular image feeds \cite{DeepSiam:triangulation} and new methods of inverse perspective mapping  \cite{DeepSiam:perspective} build up the capacity to explore 3D tracking purely based on image data. This is important in the simultaneous localization of multiple real world objects with the observation of their dynamic aspects for decision making. 
\par The primary objective of this work is to develop an online real-time multi-object tracking system for efficient human and vehicle tracking through the exploitation of appearance and spatiotemporal information through a novel Long and Short Term Memory (LSTM) \cite{DeepSiam:LSTM} based architecture along with the development of possible refinements to three dimensional track prediction through constraints observed in the Bird’s Eye View (BEV) space.

\section{Related Work}

\subsection{Single Frame Detectors}

A considerable number of new network architectures have been developed for object detection and classification where growth in accuracy and speed had been the key goals. Out of the main state-of-the-art systems, the models based on Faster-RCNN \cite{DeepSiam:FasterRCNN} that had been developed from Fast RCNN \cite{DeepSiam:FastRCNN} use a Region of Interest (RoI) to detect the objects and found to be highly accurate through the improvements with introduction of Region Proposal Network (Fully Convolutional Network that proposes regions). The single stage detectors like YOLO \cite{DeepSiam:YOLO} (You Only Look Once) network on the other hand have been optimized for speed over accuracy. They explore the entire image as a whole grid instead of computing regions of interest and can achieve high performance in frame rate (exceeding 45fps). For the industry of self driving vehicles and the real-time automated market background both accuracy and speed are crucial. A notable aspect that each architecture has adopted to improve performance is approaching localization of an anchor-based classification task followed by regression as opposed to a purely regression task. This idea will serve as one of the baselines for our work.

\subsection{Single Object Trackers}

Tracking algorithms move for deep architectures (ex: Fully Convolutional Siamese) that use deep similarity learning for tracking \cite{DeepSiam:SiamFC, DeepSiam:siammask} to solve the key challenges of changes in lighting conditions, orientation and viewpoint. Extension of these methods to the multi-object setting is yet to be achieved. Further, some algorithms have been designed to learn online to track generic objects. However, learning online needs higher computational capacity at the end device which is not a luxury that could be afforded. The work by David Held et al. on GOTURN i.e. Generic Object Tracking Using Regression Networks \cite{DeepSiam:100fps} depict the capability of achieving 100fps at test time with frozen weights. But their work is limited to single object tracking.

\subsection{Joint Tracking and Detection}

The novel idea in tracking and video recognition is the ability of improving the detection and tracking inter-dependently. That is to enhance the detection using temporal information and improve the track using both detection as well as temporal information. There has been significant progress in this area too \cite{DeepSiam:Tracktodetect, DeepSiam:mobvid}. Tracking by detection had been a considerably successful topic in the field, but this aggregation of the temporal information has turned the efforts to a different path of exploration that could predict the next level of action which is steps beyond ordinary tracking. The common idea behind these methods is the usage of temporally aware feature maps for tackling the task of detection. The key shortcoming is the lack of direct track outputs which are a requirement for tracking. 

\subsection{Multi-Object Trackers}

SORT (Simple Online and Real Time Tracking \cite{DeepSiam:Sort}) with a deep association metric \cite{DeepSiam:deepSort} presents an implementation of the Kalman filter for exploiting the temporal information and a neural network incorporating the detections and deep appearance descriptor. The key challenge faced by this work is its failure to tackle crossovers, occlusions, and modeling non-linear object motion. Improvement of the temporal aspect using the LSTMs in single object setting \cite{DeepSiam:multitarget} has presented promising results in catering to these problems. Further, the possibility of data association of random cardinality, specifically through the birth and death of characters (track initiation and termination) using LSTMs alone \cite{DeepSiam:multitarget} is equally promising. The exploitation of multiple fields of view by relating deeper layers in Siamese networks \cite{DeepSiam:multicontext} show the potential of Siamese matching even though it is considerably inhibited by the scenarios that have occlusions.

\subsection{BEV space and 3D tracking}

The methods for inverse perspective mapping and 3D detection have been extensively researched as means of depth sensation through both monocular \cite{DeepSiam:triangulation, DeepSiam:perspective} and binocular \cite{DeepSiam:triangulation} image feeds. The achievement of accuracy in depth sensation through images has approached the level of expensive range sensor data to a considerable extent. However, the task of 3D tracking is currently dominated by the algorithms that run on range sensor information \cite{DeepSiam:fastandfurious}.

\begin{figure*}[t]
\vspace{-0.7cm}
  \centering
  \includegraphics[ width=\textwidth]{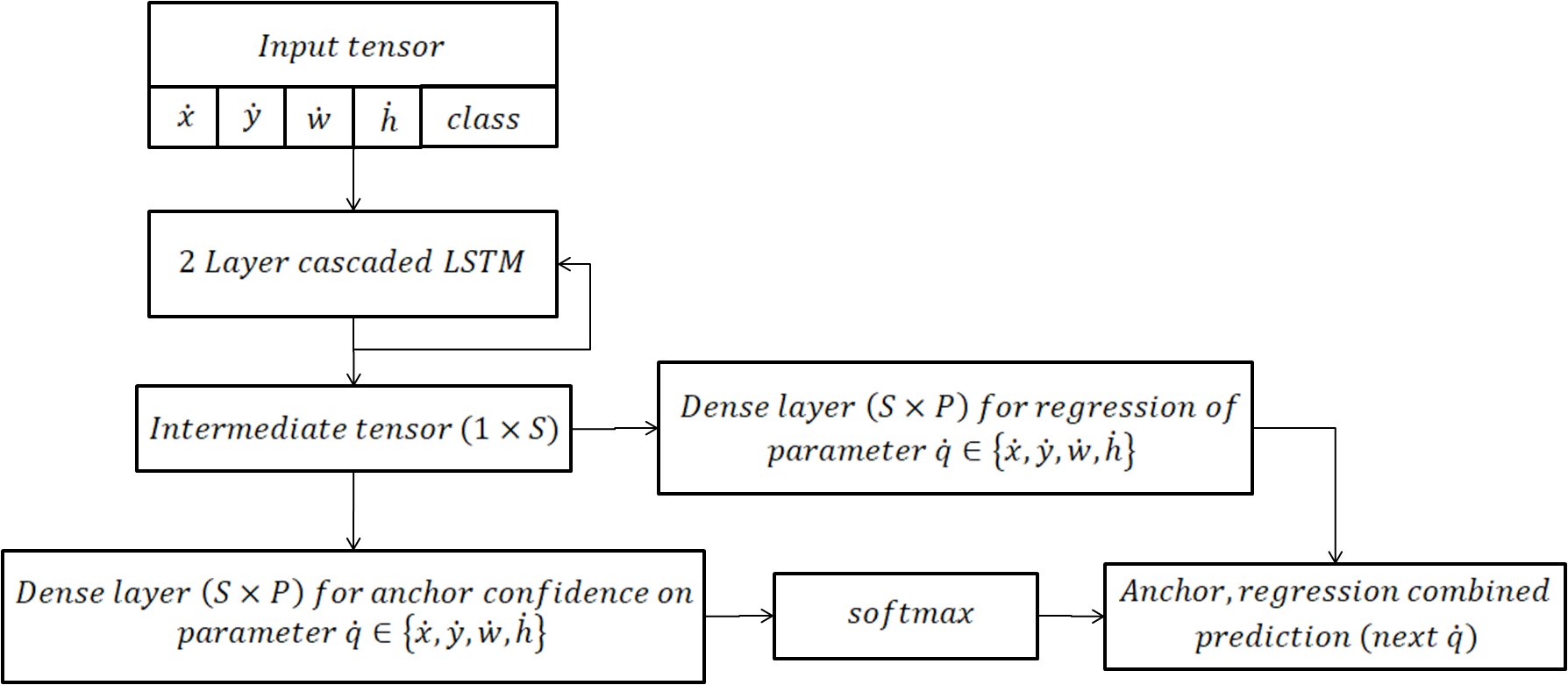}
  \vspace{-0.5cm}
  \caption{Structure of the proposed LSTM network\vspace{-0.4cm}}
  \label{fig:LSTM_network}
\vspace{-0.0cm}
\end{figure*}

\section{Methodology}

\begin{figure*}[t]
\vspace{-0.7cm}
  \centering
  \includegraphics[ width=\textwidth]{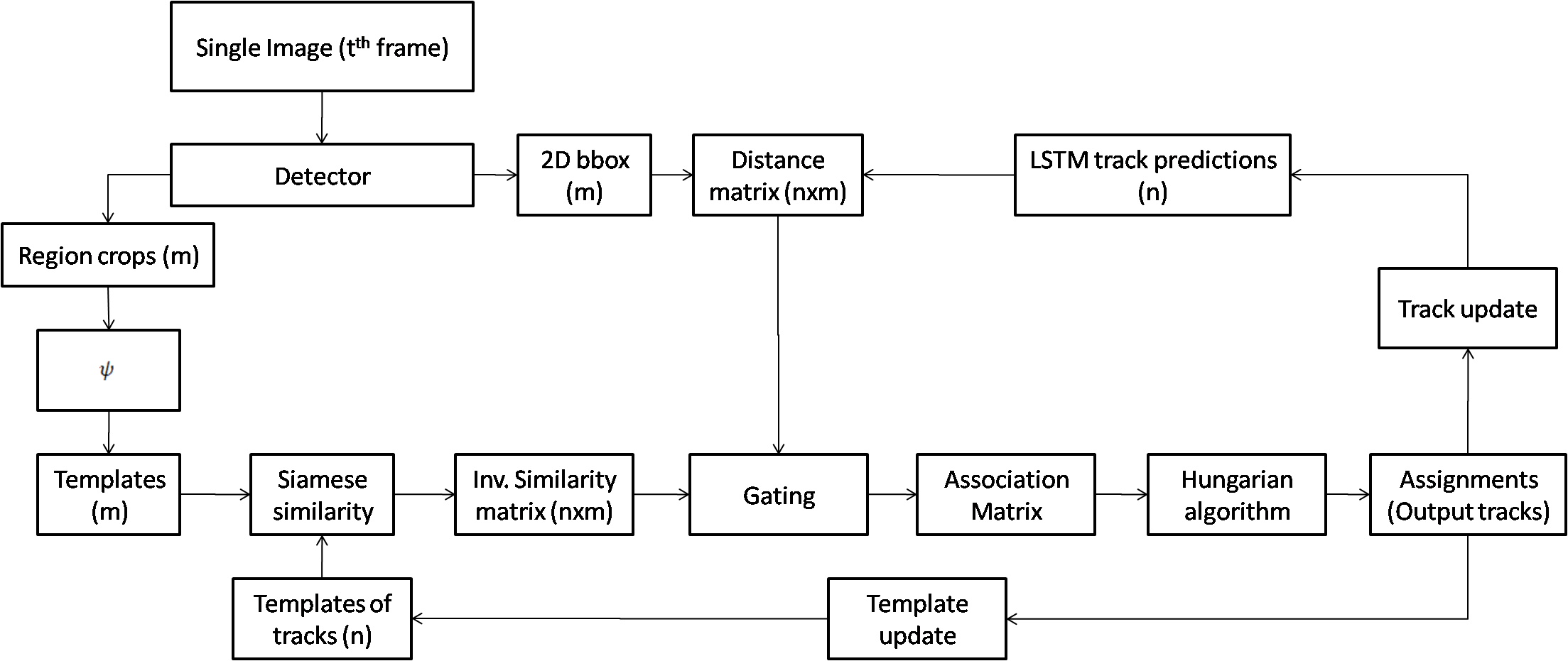}
  \vspace{-0.5cm}
  \caption{Overview of the overall 2D tracking system\vspace{-0.4cm}}
  \label{fig:tracker}
\vspace{-0.0cm}
\end{figure*}

In this section, we describe our approach for online real-time multi-object tracking. The core of our work surrounds three key attributes; an LSTM network tackling track position estimates as a probabilistic classification problem, a methodology for similarity extraction and track association that is aware of occlusions, crossovers, and other identified key challenges and finally, the extension of track predictability to the BEV space exploiting its properties. Further, we explore the possibility of propagating input uncertainties through the LSTM network. The naive integration of a generic LSTM network to exploit the temporal aspect overlooks some key aspects of the problem including uncertainty of detection positions and requirement for estimating a possible region of object presence. To address this work, we introduce an LSTM network with probabilistic outputs also capable of capturing the input uncertainties. Our final model shown in \ref{fig:tracker} performs on-par with current state-of-the-art object trackers and operates in real-time. Our model is tested on popular tracking datasets, MOT16 \cite{DeepSiam:MilanL0RS16} and KITTI-tracking \cite{DeepSiam:KITTI}. These two datasets are from slightly different domains (the former focuses on general indoor and outdoor scenes while the latter contains videos of roads taken from the perspective of a vehicle). This allows us to verify that our work solves a generalized tracking problem as opposed to a single-domain specific solution or being optimized onto a single dataset. Our work is explained under six subsections. The LSTM network along with our unique contribution is outlined initially. Then we lay out the appearance similarity usage in a multi-object domain followed by the track association problem and overall 2D tracker. Finally, we explain the extensibility of our LSTM model for accurate 3D track prediction and the improvements gained from BEV space constraints.

\subsection{LSTM network}

Consider a video as a sequence of image frames i.e. $V = I_{0},I_{1},...,I_{n}$ where $I_{k}$ s a matrix of fixed dimensions. Given detections $D = D_{0},D_{1},...,D_{n}$, for the objects present in each frame , where each  is a list of bounding box locations, class predictions, and other information corresponding to objects contained in the image , our goal here is to estimate the bounding box co-ordinates $B_{k,i}$ for each object in the following frame; $I_{k+1}$. Note that  $B_{k,i} = (x_{k,i},y_{k,i},h_{k,i},w_{k,i})$ where $x,y,h,w$ correspond to $x,y$ co-ordinates of centre, height and width of the bounding box for the $i^{th}$ object in the $k^{th}$ frame. Further, this system would operate in an online setting where at any given instance when time $t = k$, the frames, hence detections too are present only up to $I_{k}$ and $D_{k}$ espectively. Further the $i^{th}$ object will be consistent across consecutive frames (obtained using the output of the system) until the object disappears. The LSTM component can be viewed as a function $L$ with $L(D_{0},D_{1},...,D_{k}) = F_{k}$ where $F_{k}$ is a list of temporally aware feature maps $F_{k,i}$ corresponding to each object $i$ present within $D_{k}$. The remaining two functions; $C$ and $R$ correspond to classification (selecting anchor) and regression (estimating deviation from anchor) of the exact bounding box targets. Each bounding box datum $(x,y,h,w)$ is interpreted as a deviation from the previous time step $(\dot{x},\dot{y},\dot{h},\dot{w})$ which reduces the mean of those variables. Note that due to the discrete nature of data, $\dot{x} = x - x_{i-1}$ Using normalized co-ordinates $(x,y,h,w)$ values divided by relevant image dimensions); this range will be within $(-1, 1)$ and an optimum number of anchors can be used to estimate this value as a classification problem. 
\par Having laid down the classifications on to the targets, the required estimates from the classification function would be a one hot encoded tensor; $C_{out}$ of shape $(P, 4)$ for $P$ bins of anchors and 4 bounding box parameters. In our work, we use four bins; 0, 0.1, 0.5, 0.8 leading to a $(4, 4)$ tensor where the bin closest to the target value (ex: $\dot{x}$)  on each row would contain one and the rest zero. Each selected bin is an anchor located at a specific distance away from the next expected value for the parameter considered. The classification function can be presented as $C(F_{k,i}) = C_{out}$ The regression function output would be a similarly shaped tensor $R_{out}$. It is essential for the loss function to consider the nature of both the classification as well as regression outputs of the network. The overall model of estimator is illustrated in \ref{fig:LSTM_network}. Here intermediate tensor corresponds to the temporally aware feature maps $F_{k,i}$ of the $i^{th}$ object and the system has four similar but separate instances of the dense layers to handle each parameter and that finally results in outputs $C_{out}$ and $R_{out}$. In essence the network estimates how far an object would move from its current position over the next time step. The $x,y$ components capture motion along the image axes while the $h,w$ components correspond to the motion along the depth axis as well as morphological change of the object to some extent. 
\par When training; the loss function is obtained as a weighted sum of the classification and regression losses. The classification loss $LOSS_{C}$ is a simple cross-entropy loss function. The regression loss takes into account the sparse nature of the ground truth regression tensor. Here $\odot$ denotes the Hadamard product of two tensors or matrices.
$$
LOSS_{C} = -\sum(C_{out_{true}} \odot log(C_{out_{pred}})) \eqno{(1)}
$$
$$
LOSS_{R} = argmax(C_{out_{pred}}) \odot L_{Huber}(R_{out_{pred}},R_{out_{true}}) \eqno{(2)}
$$
$$
LOSS_{Total} = \lambda_{C}*LOSS_{C} + \lambda_{R}*LOSS_{R} \eqno{(3)}
$$

\subsection{Appearance similarity}

One of the most challenging problems in this context is handling occlusions. Object tracking with the use of a Kalman filter or an LSTM network to handle spatial coherence among tracks has been a common approach. However, the uncertainty involved in the track prediction increases when tracks are exposed to prolonged occlusions. Hence it is required to re-identify occluded tracks. Deep SORT \cite{DeepSiam:deepSort} introduces the use of feature vectors to define an appearance descriptor for the purpose of track re-identification. Results presented in this work have proven this to be a successful approach. However, this comes with the additional burden of training a large network for the sole purpose of re-identification of a particular class of objects. Hence this approach is not versatile for multi-class object tracking or for online implementation. Our approach has the ability to handle multiple classes of objects and can be implemented online with ease.
\par The approach implemented in this paper involves the use of a Siamese network to determine the appearance consistency of tracks. The Siamese networks described in the SiamFC \cite{DeepSiam:SiamFC, DeepSiam:endrep} and SiamMask \cite{DeepSiam:siammask} works have proven to be highly successful in single object tracking but have not been incorporated into multi-object tracking yet. It has been trained on ImageNet datasets for similarity learning and can operate online. Thus, it can give a class independent measure for appearance consistency of tracks and therefore would be ideal for track re-identification. The network discussed in SiamFC \cite{ DeepSiam:endrep} extracts the features of the exemplar image and search image to produce a cross correlation map whose peak position corresponds to the position of the object in the exemplar image within the search image. Similarly, we use a Siamese network to produce a similarity measure between two images of the same size by building up templates through a convolution neural network (a convolution function as in \cite{ DeepSiam:endrep} shown in template generation step through  in \ref{fig:tracker})
\par The cross-correlation map produced by the Siamese network is passed through a similarity function to produce the similarity score, or more accurately an appearance cost. The similarity function in this context is defined as follows,
$$
Appearance \; Cost = A \exp(-k\sum f(x,y)) \eqno{(4)}
$$
where $f(x,y)$ is the cross correlation value at $x,y$ position in the cross correlation map and $A,k$ are tunable parameters.

\subsection{Track Association}

The track association is based on the association cost which depends on the appearance cost (from the Siamese network) as well as a distance metric. The distance metric is the measurement of how far the detection bounding box is from the bounding box of a track predicted by the LSTM. The distance metric between two bounding boxes is defined based on the IOU distance (intersection over union) between the bounding boxes. Let $ a_{i,j},c_{i,j},d_{i,j}$ represent the association cost, appearance cost and the distance metric between the $i^{th}$ detection and the $j^{th}$ track.
$$ 
a_{i,j} = 
\begin{cases} 
c_{i,j} \; if \; d_{i,j} < T \\
K \; if \; d_{i,j} \geq T \\ 
\end{cases}
\eqno{(5)}
$$
where $K$ is the gating constant and $T$ is the gating threshold. Track association is treated as an assignment problem and is carried out using the Hungarian algorithm \cite{DeepSiam:hungarian} following very closely the approach discussed in Deep SORT \cite{DeepSiam:deepSort}.

\subsection{Overall online tracking system}

The Siamese network for similarity measurement is implemented in two stages. The first stage involves producing feature maps (templates) for detections in the current frame and the next stage involves producing cross correlation maps by convolving the detection templates with track templates and generating an appearance cost matrix between track, detection pairs in that frame. These two stages have been isolated to improve the efficiency of the approach.
\par In a given frame, a crop of the bounding box corresponding to each detection is extracted. These crops are resized to 127x127 and passed through the first stage of the Siamese network to generate templates for each detection in that particular frame. These templates are passed through the second stage of the Siamese network along with the templates of tracks in order to generate a matrix of appearance costs. This cost matrix is gated according to the distance metric and subjected to the Hungarian algorithm to obtain track assignments for the detections.
\par For matched track, detection pairs; the template of the track is updated using a rolling average between the track’s current template and the template of the detection which was matched to it,
$$
temp_{track} = \gamma * temp_{track} + (1 - \gamma) * temp_{det} \eqno{(6)}
$$
where $\gamma$ is the occluded percentage of the matched detection and defined as the maximum of the Intersection over area distances (IOA distances) between the detection bounding boxes and the bounding box of the matched detection.  is one when fully occluded and zero when the object is fully visible. Therefore, when the matched detection is fully visible, it replaces the template of the track with the template of the matched detection and when the matched detection is fully occluded, it does not update the template of the track so as not to contaminate the template with the features from occlusions.
\par Deletion of tracks and addition of new tracks is carried very similar to the approach carried out in the Deep SORT \cite{DeepSiam:deepSort} work.

\subsection{Extensibility to BEV space}

The seemingly simple but effective fact that ‘overlapping in BEV space projections cannot happen for the objects detected and predicted in 3D’ is exploited here through a constrained optimization problem.
\par An LSTM network is trained to predict the change of parameter ‘q’ between consecutive frames. That is, for given $\dot{q_{t-k}},...,\dot{q_{t-1}},\dot{q_{t}} \rightarrow \dot{q_{t+1}}$ is predicted where $\dot{q_{t}} = q_{t} - q_{t-1}$ and $q \in (C,S,\theta)$. Here;$C = C_{x}, C_{y}, C_{z}$ (the centre co-ordinates of the object), $S = (h,w,l)$ (object dimensions) and $\theta$ is the angle of rotation around the vertical axis.
The loss function for training the parameter predictor (LSTM) is as follows. 

\begin{multline}
LOSS_{pred}(p,\beta,\alpha,\delta,\theta) = \\
\sum_{i=1}^{N} \beta_{class_{i}} \Biggl( \Biggl( \sum_{p\in (C,S)}\alpha_{p}L_{Huber,\delta_{p}}(p_{pred},p_{gt}) \Biggr) + \\ \alpha_{\theta}L_{\theta}(\theta_{pred},\theta_{gt})_{object=i} \Biggr)
\tag{7}
\end{multline}

Here $P_{pred}$ refers to the predicted parameter and $P_{gt}$ refers to the ground truth parameter. 
\par $\delta_{p}$ is a parameter based learnable which in turn is the quadratic-linear margin of the Huber loss function and $\alpha_{p}$ or $\alpha_{\theta}$ is a regressed parameter based learnable (where in the case of $\alpha_{\theta}$, the regressed parameter is $\theta$ and $\alpha_{p}$ is similarly interpreted whereas the scope of $\alpha_{p}$ is different from that of $\delta_{p}$, considering the impact on cost function) and $\beta_{class i}$ is the class based learnable parameter w.r.t. the class of the $i^{th}$ object.
\par Here, $p = C_{x},C_{y},C_{z},h,w,l$ , $\beta = \beta_{class} | class\in classes$, $\alpha=[[\alpha_{p}]_{p \in parameters}, \alpha_{\theta}]$ and $\delta=[\delta_{p}]_{p \in parameters}$.
\par Due to the discontinuous nature of the parameter $\theta$ at the two extreme ends of its domain $[-\pi, \pi]$, and due to the fact that $\theta = \pi$ and $\theta = -\pi$ depict the same orientation, it is not directly incorporated into the Huber loss function. It is handled separately using $L_{\theta}$ function \cite{DeepSiam:geometry}, where $\theta_{pred},\theta_{gt}$ are predicted and ground truth values of the parameter $\theta$ respectively.
$$
L_{\theta}(\theta_{pred},\theta_{gt}) = 0.5(1 - \cos(\theta_{gt} - \theta_{pred})) \eqno{(8)}
$$

\subsection{Constraints as penalties}
First, we introduce the hard constraint on BEV space that projections of the objects on to the x-z plane in general co-ordinates have no intersection. However, most of the research is focused on building up 3D bounding boxes of objects where the rectangular projection does not create a clear cut segmentation of the object (ex: human) on BEV space. Therefore, we minimize an additional term as follows.
$$
I = \sum_{v_{i},v_{j} \in objects_{pred}, i \neq j} (1 + \xi^{2}_{class_{i}, class_{j}})(v_{i_{BEV}} \cap v_{j_{BEV}})  \eqno{(9)}
$$
Where $v_{i_{BEV}}$ is the projection of the bounding box of the object $v_{i}$ onto the BEV space and $\xi_{class_{i},class{j}}$ is a learnable based on object classes under intersection which in turn forms a set $\xi_{class \times class}$ and each term is squared to ensure positivity.
Therefore, the final minimization function is as follows,
$$
L(p,\beta, \alpha, \delta, \theta, \{\xi\}) = LOSS_{pred}(p,\beta, \alpha, \delta, \theta) + I \eqno{(10)}
$$
However, at an optimum point $(p^{*},\beta^{*}, \alpha^{*}, \delta^{*}, \theta^{*}, \{\xi\}^{*})$; the loss function obeys a feature observed in Lagrange constrained optimization that;
$\nabla L = 0$ where $\nabla$ refers to the discrete derivative (this statement is intuitive only with the discrete derivative).
\par This implies that:
$$
\nabla_{p,\theta}Loss_{pred} = -(1 + \xi^{2}_{class_{i}, class_{j}})\nabla_{p,\theta}(v_{i_{BEV}} \cap v_{j_{BEV}})  \eqno{(11)}
$$
for all classes at optimum parameters $p^{*},\theta^{*}$. Therefore $(1 + \xi^{2}_{class_{i}, class_{j}})$ behaves similar to a Lagrange multiplier. This setting helps to build up a network that trains not only based on the individual performance per object but also encountering the joint effect of multiple object scenarios.

\begin{table*}[h]
\caption{Comparison  of  our  performance  on  MOT16  dataset  with  recent  works}
\label{MOT16 Results}
\begin{center}
\begin{tabular}{|c||c||c||c||c||c|}
\hline
Method & Mode & MOTA$\uparrow$ & MOTP$\uparrow$ & MT$\uparrow$ & ML$\downarrow$\\
\hline
Deep SORT \cite{DeepSiam:deepSort} & ONLINE & 61.40\% & 79.10\% & 32.80\% & 18.20\%\\
\hline
SORT \cite{DeepSiam:Sort} & ONLINE & 59.80\% & 79.60\% & 25.40\% & 22.70\%\\
\hline
RNN LSTM \cite{DeepSiam:multitarget} & ONLINE & 19.00\% & 71.00\% & 05.50\% & 45.60\%\\
\hline
MDP \cite{DeepSiam:learntotrack} & ONLINE & 30.30\% & 71.30\% & 13.00\% & 38.40\%\\
\hline
DMAN \cite{DeepSiam:attention} & ONLINE & 46.10\% & 73.80\% & 17.40\% & 42.70\%\\
\hline
LSTM+Similarity (Ours) & ONLINE & 66.70\% & 69.00\% & 39.18\% & 16.80\%\\
\hline
Kalman Filter (Ours) & ONLINE & 61.00\% & 69.00\% & 17.00\% & 17.00\%\\
\hline
\end{tabular}
\end{center}
\end{table*}

\begin{table*}[h]
\caption{Comparison  of  our  performance  on  KITTI-trracking  dataset  with  recent  works}
\label{KITTI Results}
\begin{center}
\begin{tabular}{|c||c||c||c||c||c|}
\hline
Method & Mode & MOTA$\uparrow$ & MOTP$\uparrow$ & MT$\uparrow$ & ML$\downarrow$\\
\hline
Regionlets Only \cite{DeepSiam:kalman} & ONLINE & 76.40\% & 81.50\% & 54.10\% & 9.30\%\\
\hline
MS-CNN Only \cite{DeepSiam:kalman} & ONLINE & 81.23\% & 85.60\% & 66.30\% & 4.60\%\\
\hline
Regionlets MS-CNN \cite{DeepSiam:kalman} & ONLINE & 82.60\% & 85.00\% & 70.50\% & 5.30\%\\
\hline
SMES \cite{DeepSiam:simmap} & ONLINE & 70.78\% & 80.38\% & 51.68\% & 7.77\%\\
\hline
LSTM + Similarity (Ours) & ONLINE & 83.58\% & 78.50\% & 48.23\% & 2.25\%\\
\hline
\end{tabular}
\end{center}
\end{table*}

\section{Results}

\subsection{Datasets and Evaluation metrics}
Experiments are conducted on the MOT16 \cite{DeepSiam:MilanL0RS16} and KITTI \cite{DeepSiam:KITTI} tracking datasets. The MOT16 dataset contains 7 videos in its training set. The KITTI tracking dataset contains 21 videos in its training set. The Siamese Network for appearance consistency is trained completely on external data (ImageNet datasets) and there is no overlap with any of the MOT16 or KITTI data. The LSTM network is trained only with the use of bounding box locations of objects and class information for a partition of the training sets of these two datasets (the remainder is kept aside for testing purposes). Results are reported for our test partition (in the case of LSTM usage) and for the entire datasets (in cases they are not used for training).
\par Evaluation of our system is carried out for the entire system as well as for the study of LSTM network alone. For the case of the entire system, we consider the metrics used by the MOT benchmarks for evaluation. This includes Multiple Object Tracking Accuracy (MOTA), Multiple Object Tracking Precision (MOTP), the ratio of Mostly Tracked targets (MT), and the ratio of Mostly Lost targets (ML). In the case of the LSTM network, the Average Precision (AP) value for the predicted frames across the dataset and classes is reported.

\subsection{Evaluation}
The evaluations on the MOT16 Dataset for the end to end system are reported in Table I. Evaluations mainly focus on two aspects: improvement in accuracy with the introduction of the similarity measure to a traditional tracker using only a Kalman filter or an LSTM network and how closely related the accuracy is with state of the art multi-object trackers. Similar results on the KITTI tracking dataset are presented for our work alongside comparisons (note that few state-of-the-art works report on this dataset) in Table II. Separate evaluations for the LSTM in the case of single object tracking for individual tracklets in the KITTI dataset was carried out. An average IoU of 61.45 and AP of 0.96 at 0.5 IoU were obtained for this experiment.

\section{Conclusion and future work}

We present an end-to-end system capable of performing multi-object tracking by combining a range of advances in object detection and reidentification along with our novel architectures and loss functions. Further, we work on a novel step by building a separate LSTM branch to estimate the similarity feature map for the next time step of a given track. The Siamese Networks may be viewed as a two-step version of our extension, whereas this replacement with an LSTM is more of a generalized version capable of generating a better feature set. The key expectation with this addition is the overcoming of identity switches and lost tracks in the case of occlusions. Appearance features tend to change significantly during an occlusion, especially when an object undergoes rotations, and our extension overcomes this by modeling the appearance changing pattern over time. 

\addtolength{\textheight}{-12cm}


\bibliographystyle{IEEEtran}
\bibliography{DeepSiam}

\begin{thebibliography}{10}
\providecommand{\url}[1]{#1}
\csname url@rmstyle\endcsname
\providecommand{\newblock}{\relax}
\providecommand{\bibinfo}[2]{#2}
\providecommand\BIBentrySTDinterwordspacing{\spaceskip=0pt\relax}
\providecommand\BIBentryALTinterwordstretchfactor{4}
\providecommand\BIBentryALTinterwordspacing{\spaceskip=\fontdimen2\font plus
\BIBentryALTinterwordstretchfactor\fontdimen3\font minus
  \fontdimen4\font\relax}
\providecommand\BIBforeignlanguage[2]{{%
\expandafter\ifx\csname l@#1\endcsname\relax
\typeout{** WARNING: IEEEtran.bst: No hyphenation pattern has been}%
\typeout{** loaded for the language `#1'. Using the pattern for}%
\typeout{** the default language instead.}%
\else
\language=\csname l@#1\endcsname
\fi
#2}}

\bibitem{DeepSiam:SiamFC}
L.~Bertinetto, J.~Valmadre, J.~F. Henriques, A.~Vedaldi, and P.~H.~S. Torr,
  ``{Fully-convolutional siamese networks for object tracking},'' in
  \emph{European Conference on Computer Vision (ECCV)}, 2016.

\bibitem{DeepSiam:siammask}
L.~Z. Qiang~Wang, L.~Bertinetto, W.~Hu, and P.~H.~S. Torr, ``{Fast online
  object tracking and segmentation: A unifying approach},'' in \emph{Conference
  on Computer Vision and Pattern Recognition (CVPR)}, 2019.

\bibitem{DeepSiam:kalman}
J.~Kuck and P.~Zhuang, \emph{Target tracking with kalman filtering}, 2016.

\bibitem{DeepSiam:multitarget}
A.~Milan, S.~H. Rezatofighi, A.~Dick, I.~Reid, and K.~Schindler, ``{Online
  multi-target tracking using recurrent neural networks},'' in \emph{Conference
  on Artificial Intelligence, Association for the Advancement of Artificial
  Intelligence (AAAI)}, 2016.

\bibitem{DeepSiam:deeptracking}
P.~Ondruska and I.~Posner, ``{Deep tracking: Seeing beyond seeing using
  recurrent neural networks},'' in \emph{Conference on Artificial Intelligence,
  Association for the Advancement of Artificial Intelligence (AAAI)}, 2016.

\bibitem{DeepSiam:deepSort}
N.~Wojke, A.~Bewley, and D.~Paulus, ``{Simple online and realLong Short-term
  Memorytime tracking with a deep association metric},'' in \emph{IEEE
  International Conference on Image Processing (ICIP)}, 2017.

\bibitem{DeepSiam:triangulation}
Z.~Qin, J.~Wang, and Y.~Lu, ``{Triangulation Learning Network: from Monocular
  to Stereo 3D Object Detection},'' in \emph{Conference on Computer Vision and
  Pattern Recognition (CVPR)}, 2019.

\bibitem{DeepSiam:perspective}
T.~Bruls, HoriaPorav, L.~Kunze, and P.~Newman, ``{The Right (Angled)
  Perspective: Improving the Understanding of Road Scenes Using Boosted Inverse
  Perspective Mapping},'' in \emph{IEEE Intelligent Vehicles Symposium}, 2019.

\bibitem{DeepSiam:LSTM}
H.~S. and S.~J., ``Long short-term memory,'' \emph{Neural Comput.}, vol.~9,
  no.~8, pp. 1735--1780, 1997.

\bibitem{DeepSiam:FasterRCNN}
S.~Ren, K.~He, R.~B. Girshick, and J.~Sun, ``{Faster R-CNN: towards real-time
  object detection with region proposal networks},'' in \emph{IEEE Transactions
  on Pattern Analysis and Machine Intelligence}, 2015.

\bibitem{DeepSiam:FastRCNN}
R.~B. Girshick, ``{Fast R-CNN},'' in \emph{International Conference on Computer
  Vision (ICCV)}, 2015.

\bibitem{DeepSiam:YOLO}
J.~Redmon, S.~Divvala, R.~Girshick, and A.~Farhadi, ``{You only look once:
  Unified, real-time object detection},'' in \emph{Conference on Computer
  Vision and Pattern Recognition (CVPR)}, 2016.

\bibitem{DeepSiam:100fps}
D.~Held, S.~Thrun, and S.~Savarese, ``{Learning to track at 100 FPS with deep
  regression networks},'' in \emph{European Conference on Computer Vision
  (ECCV)}, 2016.

\bibitem{DeepSiam:Tracktodetect}
C.~Feichtenhofer, A.~Pinz, and A.~Zisserman, ``{Detect to track and track to
  detect},'' in \emph{International Conference on Computer Vision (ICCV)},
  2017.

\bibitem{DeepSiam:mobvid}
M.~Liu and M.~Zhu, ``{Mobile video object detection with temporally-aware
  feature maps},'' in \emph{Conference on Computer Vision and Pattern
  Recognition (CVPR)}, 2018.

\bibitem{DeepSiam:Sort}
A.~Bewley, Z.~Ge, L.~Ott, F.~Ramos, and B.~Upcroft, ``{Simple online and
  realtime tracking},'' in \emph{IEEE International Conference on Image
  Processing (ICIP)}, 2016, p. 3464–3468.

\bibitem{DeepSiam:multicontext}
H.~Morimitsu, ``{Multiple Context Features in Siamese Networks for Visual
  Object Tracking},'' in \emph{European Conference on Computer Vision (ECCV)},
  2018.

\bibitem{DeepSiam:fastandfurious}
W.~Luo, B.~Yang, and R.~Urtasun, ``{Fast and Furious: Real Time End-to-End 3D
  Detection, Tracking and Motion Forecasting with a Single Convolutional
  Net},'' in \emph{Conference on Computer Vision and Pattern Recognition
  (CVPR)}, 2018.

\bibitem{DeepSiam:MilanL0RS16}
\BIBentryALTinterwordspacing
A.~Milan, L.~Leal{-}Taix{\'{e}}, I.~D. Reid, S.~Roth, and K.~Schindler,
  ``{MOT16:} {A} benchmark for multi-object tracking,'' \emph{CoRR}, vol.
  abs/1603.00831, 2016. [Online]. Available:
  \url{http://arxiv.org/abs/1603.00831}
\BIBentrySTDinterwordspacing

\bibitem{DeepSiam:KITTI}
A.~Geiger, P.~Lenz, and R.~Urtasun, ``{Are we ready for autonomous driving? the
  kitti vision benchmark suite},'' in \emph{Conference on Computer Vision and
  Pattern Recognition (CVPR)}, 2012.

\bibitem{DeepSiam:endrep}
J.~Valmadre, L.~Bertinetto, J.~Henriques, A.~Vedaldi, and P.~H.~S. Torr,
  ``{End-to-end representation learning for correlation filter based
  tracking},'' in \emph{Conference on Computer Vision and Pattern Recognition
  (CVPR)}, 2017, p. 5000–5008.

\bibitem{DeepSiam:hungarian}
H.~W. Kuhn, ``The hungarian method for the assignment problem,'' \emph{Naval
  Research Logistics (NRL)}, 1955.

\bibitem{DeepSiam:geometry}
J.~F. Arsalan~Mousavian, Dragomir~Anguelov, ``{3D Bounding Box Estimation Using
  Deep Learning and Geometry},'' in \emph{Conference on Computer Vision and
  Pattern Recognition (CVPR)}, 2017.

\bibitem{DeepSiam:learntotrack}
Y.~Xiang, A.~Alahi, and S.~Savarese, ``{Learning to track: Online multi-object
  tracking by decision making},'' in \emph{International Conference on Computer
  Vision (ICCV)}, 2015, p. 4705–4713.

\bibitem{DeepSiam:attention}
J.~Zhu, H.~Yang, N.~Liu, M.~Kim, W.~Zhang, and M.-H. Yang, ``Online
  multi-object tracking with dual matching attention networks,'' \emph{Computer
  Vision – ECCV}, vol. 11209, p. 379–396, 2018.

\bibitem{DeepSiam:simmap}
M.~Kim, S.~Alletto, and L.~Rigazio, ``{Similarity mapping with enhanced siamese
  network for multi-object tracking},'' in \emph{Machine Learning for
  Intelligent Transportation Systems (MLITS)}, 2016.

\end{thebibliography}

\end{document}